\pdfoutput=1

\PassOptionsToPackage{table}{xcolor}
\documentclass[11pt]{article}

\usepackage[]{ACL2023}

\usepackage{times}
\usepackage{latexsym}
\usepackage{hwemoji}
\usepackage[T1]{fontenc}

\usepackage[utf8]{inputenc}
\usepackage{microtype}

\usepackage{inconsolata}

\usepackage{enumerate}
\usepackage{enumitem}
\usepackage{color}
\usepackage{adjustbox}
\usepackage{booktabs}
\usepackage{makecell}
\usepackage{amsfonts}
\usepackage{amsthm}
\usepackage{amssymb}
\usepackage{amsmath}
\usepackage{comment}
\usepackage{multirow}
\usepackage{xfrac}
\usepackage{scalerel}
\usepackage{float}
\usepackage{inconsolata} 
\usepackage[belowskip=-10pt,aboveskip=10pt]{caption}
\usepackage{cleveref}
\usepackage{tipa}
\usepackage{graphicx,xparse}
\usepackage{rotating}
\usepackage{pdflscape}
\usepackage{titlesec}
\titlespacing*{\section}{0pt}{0.7\baselineskip}{0.5\baselineskip}

\crefformat{section}{\S#2#1#3} 
\crefformat{subsection}{\S#2#1#3}
\crefformat{subsubsection}{\S#2#1#3}

\newcommand{\deltaLL}{{\Delta_\mathrm{LL}}}
\newcommand{\xint}{x^q}

\newcommand{\ie}{\emph{i.e., }}

\newcommand{\pp}{\mathrm{\Delta\mathrm{PP}}}

\newcommand{\defeq}[0]{\mathrel{\stackrel{\textnormal{\tiny def}}{=}}}
\newcommand{\db}[1]{\textcolor{darkblue}{\bf\scriptstyle \selectfont \,(\pm#1)}}

\NewDocumentCommand\uzhem{}{\scalerel*{\includegraphics{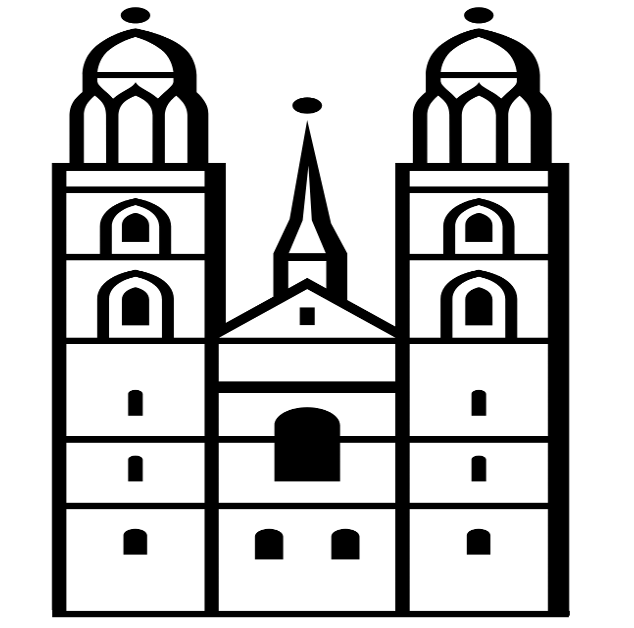}}{X}}
\NewDocumentCommand\pdem{}{\scalerel*{\includegraphics{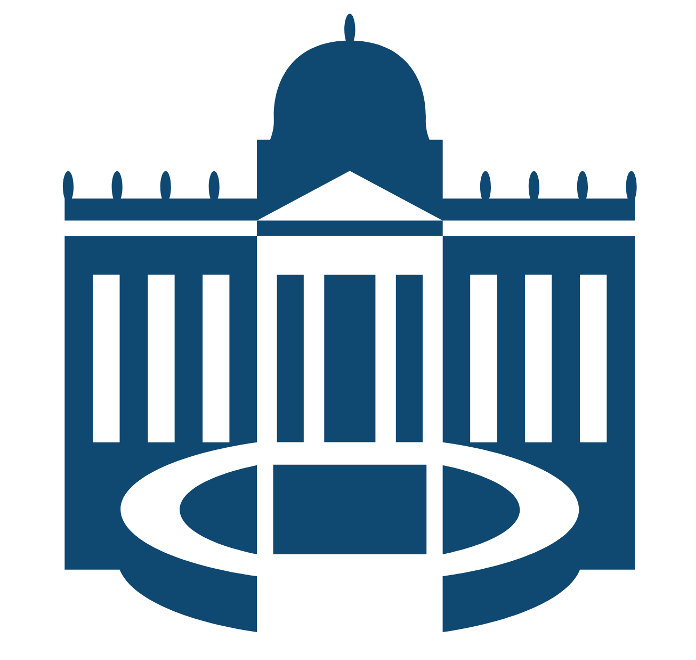}}{X}}

\title{Language models emulate certain cognitive profiles: An investigation of how predictability measures interact with individual differences}

\newcommand{\uzh}{\scaleobj{1.6}{\uzhem}}
\newcommand{\up}{{\scaleobj{1.6}\pdem}}

\author{~\;~Patrick Haller$^{\uzh}$,~\;~Lena S. Bolliger$^{\uzh}$,~\;~Lena A. Jäger$^{\uzh,\up}$\\
$^{\uzh}$Department of Computational Linguistics, University of Zurich, Switzerland\\$^{\up}$Department of Computer Science, University of Potsdam, Germany\\
\texttt{\{\href{mailto:haller@cl.uzh.ch}{haller},\href{mailto:bolliger@cl.uzh.ch}{bolliger},\href{mailto:jaeger@cl.uzh.ch}{jaeger}\}@cl.uzh.ch}}

\begin{document}
\maketitle
\begin{abstract}
To date, most investigations on surprisal and entropy effects in reading have been conducted on the group level, disregarding individual differences. In this work, we revisit the predictive power of surprisal and entropy measures estimated from a range of language models (LMs) on data of human reading times as a measure of processing effort by incorporating information of language users' cognitive capacities. To do so, we assess the predictive power of surprisal and entropy estimated from generative LMs on reading data obtained from individuals who also completed a wide range of psychometric tests.
Specifically, we investigate if modulating surprisal and entropy relative to cognitive scores increases prediction accuracy of reading times, and we examine whether LMs exhibit systematic biases in the prediction of reading times for cognitively high- or low-performing groups, revealing what type of psycholinguistic subject a given LM emulates.
Our study finds that in most cases, incorporating cognitive capacities increases predictive power of surprisal and entropy on reading times, and that generally, high performance in the psychometric tests is associated with lower sensitivity to predictability effects. Finally, our results suggest that the analyzed LMs emulate readers with lower verbal intelligence, suggesting that for a given target group (i.e., individuals with high verbal intelligence), these LMs provide less accurate predictability estimates.\footnote{
Code is available at \url{https://github.com/DiLi-Lab/LM-cog-profiles} 
}
\end{abstract}

\section{Introduction}
\begin{figure}[htb!]
    \centering
    \includegraphics[width=.9\linewidth]{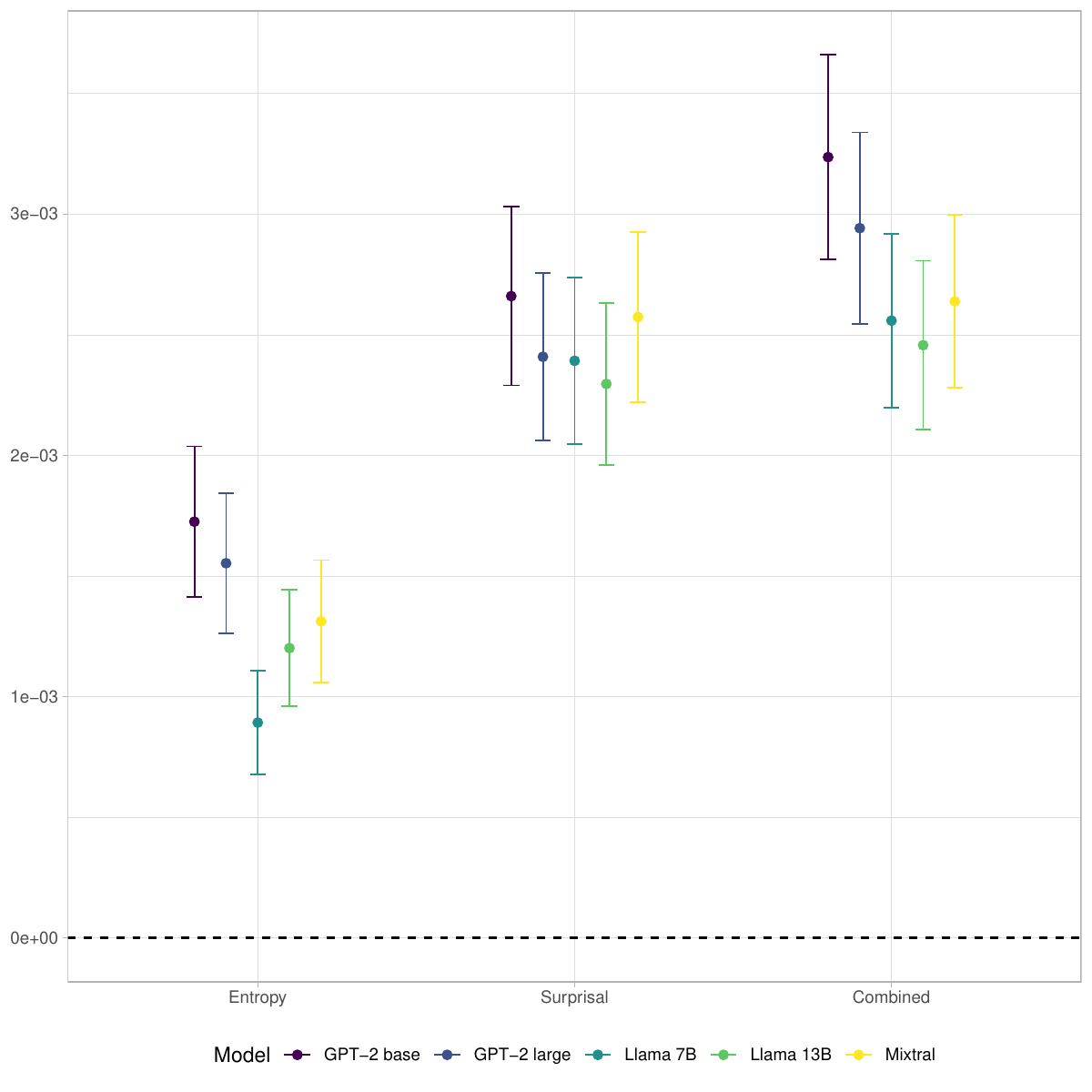}
    \caption{Predictive power of entropy and surprisal on reading times. Combined refers to the regression model where both predictors were included. Higher $\deltaLL$ indicates higher predictive power. \looseness=-1
    } \label{fig:baseline}
    \setlength{\belowcaptionskip}{-30pt}
\end{figure}
Human language comprehension and, by extension, human reading is incremental in nature: humans process words sequentially~\citep{rayner2009language}, and different words in varying contexts impose different amounts of cognitive processing effort~\citep{rayner:1998}.
Similarly, language models' conditional probability distributions assign different probabilities for potential continuations for a given prefix. The relationship between cognitive effort and predictability measures derived from LMs' probability distribution was operationalized by \emph{surprisal theory}~\citep{hale2001probabilistic, levy2008expectation}.
Since then, a large body of research has investigated the exact nature of the relationship between surprisal and human processing effort, such as determining appropriate linking functions~\citep{meister-etal-2021-revisiting, shain2024large}, or its manifestation in different languages~\citep[][i.a.]{wilcox-etal-2023-language, jager2015subject,kuribayashi-etal-2021-lower}. Moreover, it has been shown repeatedly that both the quality of a model from which surprisal is extracted as well as the amount of data a model is trained on correlate with the model's psychometric predictive power
\footnote{This hypothesis has been referred to as \emph{quality-power hypothesis} by \citet{wilcox-etal-2023-language}, cf.~their introduction for a comprehensive summary.}, \ie its ability to predict human behavioral processing data~\citep{frank2011insensitivity, fossum2012sequential, goodkind2018predictive}, albeit only to a certain extent~\citep{shain2024large, oh2023does, oh-schuler-2023-transformer}. 
So far, most studies have tested these predictions on the group-level, neglecting individual cognitive differences that might influence readers' capacities to make predictions about upcoming material. However, an increasing body of research has demonstrated that an individual's characteristics do play an important role in human language processing \citep[e.g.,][]{Estes1956,Daneman1980,Levinson2012,VanDyke2014}. 
Apart from surprisal, it has been suggested that reading processes are not purely \emph{responsive}, which is mirrored in the surprisal effect of a word on a reader, but also \emph{anticipatory}: readers make implicit assumptions about future words and allocate time to process it in advance. This anticipatory effect, which is reflected in the reading behavior, is induced by a reader's \emph{expectation} about a word's surprisal and is operationalized as a word's \emph{contextual entropy}~\citep{linzen2016uncertainty, van2017approximations, cevoli2022prediction, pimentel2023effect}.

In this work, we revisit the relationship between predictability measures (surprisal and contextual entropy) and data of human processing effort in consideration of language users' individual cognitive differences. More specifically, we assess the \emph{predictive power} (PP) of entropy, surprisal, as well as their interactions with cognitive measures, as predictors of human reading times in linear-mixed models. After establishing the baseline predictive power of surprisal and entropy on our German reading data (\textbf{H}\textsubscript{B}), we investigate the following novel hypotheses:

\begin{enumerate}[itemsep=0pt, parsep=0pt]
    \item[\textbf{H}$_1$:] Modulating surprisal and entropy effects relative to individual cognitive capacities improves their predictive power on reading times on unseen data. 
    \item[\textbf{H}$_2$:] Individuals with higher cognitive capacity rely less on predictive processing strategies, and hence exhibit lower surprisal or entropy effects. 
    \item[\textbf{H}$_3$:] LMs are better at predicting reading times for certain cognitive profiles. 
\end{enumerate}


To address these hypotheses, we utilize the \emph{Individual Differences Corpus}~\citep[InDiCo;][]{haller2023measurement}, which contains both reading data and scores of a comprehensive psychometric assessment targeting various cognitive capacities, including verbal and non-verbal working memory, verbal and non-verbal cognitive control, verbal and non-verbal intelligence and reading fluency. We deploy five pre-trained generative LMs from three language-families---GPT-2 base and large, Llama 2 7B and 13B, and Mixtral---to estimate both surprisal and contextual entropy and quantify their predictive power by including them as predictors in linear regressors, which are fitted to predict by-word reading times from InDiCo. If the regressors' log-likelihood improves after including these predictors and their interaction with the psychometric scores, we take this as corroboration of their predictive power.

We find that adding interaction terms between predictability measures (surprisal and entropy) and most cognitive scores significantly improves the quality of reading time predictions, and that in general, individuals with higher cognitive capacities exhibit smaller predictability effects. Lastly, there is evidence that LMs' abilities to predict reading times vary between high- and low-performing individuals within certain cognitive capacities. Specifically, all tested models emulate the processing behaviour of individuals with low verbal intelligence.

Our work is a first step towards investigating i) the differences in surprisal and entropy effects across different cognitive profiles, and ii) what type of cognitive biases might be inherent in the way LMs process language.




\section{Related work}
\label{sec:rel-work}

\subsection{Surprisal and predictive power}
\label{sec:rel-work-surprisal}
Surprisal is a measure of predictability of a word in its context and has shown to be proportional to cognitive effort in human sentence processing~\citep{hale2001probabilistic, levy2008expectation}. It is quantified as the negative log probability of a word given its preceding context. 
Since the formalization of surprisal theory, many studies have corroborated its linear relationship with reading times~\citep{demberg2008data, shain2021cdrnn, hoover2023plausibility, pimentel2023effect}, not just in English but also across languages~\citep{pimentel+al.emnlp2021a, wilcox-etal-2023-language, wilcox2023testing, de-varda-marelli-2022-effects, jager2015subject,kuribayashi-etal-2021-lower}. Moreover, researchers have investigated the degree to which surprisal is predictive of human reading times (i.e., assessing the \emph{predictive power} (PP) of surprisal on human reading times) deploying different LMs. \citet{wilcox2020predictive} found that the better an LM's next-word expectation (\ie the lower its perplexity), the higher its PP. Along the same line, \citet{wilcox-etal-2023-language} demonstrated that an increasing LM quality, quantified by decreasing cross-entropy during training, leads to surprisal values that better predict RTs. Similarly, \citet{goodkind-bicknell:2018predictive} showed that the PP of surprisal increases linearly with the quality of LMs. This finding has since been refuted by \citet{oh2023does} who revealed that large models, despite lower perplexity, provide worse PP of RTs. \citet{oh-schuler-2023-transformer} further demonstrated that LMs provide the best fit to RTs when trained on around 2 billion tokens; beyond that point, additional training data causes the PP to decrease again. 

\subsection{Contextual entropy and predictive power}
\label{sec:rel-work-entropy}
\citet{linzen2016uncertainty} examined how sentence processing is affected by readers' uncertainty about the predictions they make during processing. They found that what they term single-step entropy (contextual entropy, cf.\cref{sec:methods:ce}) does not affect RTs. However, they computed single-step entropy only over upcoming constituents based on verb subcategorization frames; it was later shown that entropy is indeed predictive of RTs~\citep{van2017approximations}. \citet{cevoli2022prediction} looked at the interaction of surprisal and entropy when predicting RTs: the impact of surprisal on reading behavior should vary as a function of entropy, such that surprising words inflict particularly high processing load when entropy is low. \citet{wilcox2023testing} examined whether contextual entropy is predictive of reading times and discovered that adding entropy as additional predictor (while keeping surprisal) increases PP, while replacing surprisal with entropy leads to a decrease in PP. However, \citet{pimentel2023effect} also showed that using contextual entropy as a predictor in a linear-mixed model can be as good as surprisal when analyzing \emph{anticipatory} effects reflected in word skipping rates, as opposed to \emph{responsive} effects captured by gaze duration, for instance.

\subsection{Individual differences in sentence processing}
\label{sec:rel-work-individual-differences}

Theories of sentence processing generally assume that the cognitive mechanisms involved in language processing are qualitatively identical across speakers. 
However, this perspective has been challenged, with evidence emerging that differences in cognitive abilities among language users do indeed have a significant impact on processing \citep[i.a.]{Vuong2014,Nicenboim2015,Farmer2017}. For instance, \citet{Kuperman2011} demonstrated that measures related to cognitive control interact with word length and lexical frequency effects on fixation times, and \citet{Nicenboim2015} showed that readers ranking lower in working-memory tests exhibit more regressive saccades in regions with high memory load.

Several studies have also investigated individual differences in surprisal effects, in particular in the realm of native and non-native reading~\citep{berzak2023eye, schneider2023non}. For instance, \citet{berzak2023eye} demonstrated that higher L2 proficiency is associated with increased sensitivity to a word’s predictability in context (surprisal). Moreover, \citet{vskrjanec2023expert} showed that specialized surprisal from domain-adapted LMs improves reading-time predictions for expert readers.

\section{Methods}

\paragraph{Surprisal.}
Given a vocabulary $\Sigma$ and an augmented vocabulary $\Bar{\Sigma} = \Sigma \cup \{\textsc{eos}\}$, which contains a special \textsc{eos} (end-of-sentence) token, the surprisal \citep{shannon1948mathematical} of a given sequence is defined as 
\begin{equation}
    s(u_n) \defeq - \log p(u_n \mid \mathbf u_{< n}),
\end{equation} where $p(\cdot \mid \mathbf{u}_{<n})$ is the true distribution over words $u \in \Bar{\Sigma}$ in context $\mathbf{u}_{<n}$. In other words, surprisal of a word is the negative log-probability conditioned on its left context.

\paragraph{Contextual entropy.} \label{sec:methods:ce}
The contextual entropy of a $\Bar{\Sigma}$-valued random variable $U_n$ at index $n$ is the expected value of its surprisal, formalized as 
\begin{equation}
\small
    \begin{split}
        \mathrm{H}(U_n \mid \mathbf{U}_{<n} = \mathbf{u}_{<n})
      \defeq \mathbb{E}_{u \sim p(\cdot|\mathbf{u}_{<n})}\left[s_n(u)\right] \\ = 
        -\sum_{u \in \Bar{\Sigma}} p(u | \mathbf{u}_{<n}) \log_2 p(u | \mathbf{u}_{<n}).
    \end{split}
\end{equation}
It is a specific version of the Shannon entropy $\mathrm{H}(U) \defeq -\sum_{u \in \mathcal{U}} p(u) \log p(u)$ that is conditioned on the left context of $U$~\citep{shannon1948mathematical}.
As we do not have access to the true distribution $p(\cdot \mid \mathbf{u}_{<n})$, we approximate both measures using an auto-regressive language model $p_{\theta}$.


\subsection{Assessing predictive power}
We utilize linear-mixed models (LMMs) $\mathcal{M}$ to predict a reading time measure $y_{ij}$, obtained from a subject $j$ on word $i$, from a set of standardized word-level and subject-level predictors $\mathbf{x}_{ij}$, \ie $\mathcal{M}: \mathbf{x}_{ij} \mapsto y_{ij}$. 

For our analyses, we want to quantify the predictive power of a given predictor of interest $\xint$ (\emph{e.g.,} surprisal). To do so, we first define a baseline model $\mathcal{M}^b: \mathbf{x}^b_{ij} \mapsto y_{ij}$ that includes a set of baseline predictors $\mathbf{x}^b_{ij}$, and a target model $\mathcal{M}^t: \mathbf{x}^b_{ij} \oplus\xint_{ij} \mapsto y_{ij}$ that additionally includes the predictor of interest $\xint_{ij}$, where $\oplus$ represents the concatenation of two sets of predictors. Following previous work \citep[i.a.]{wilcox2020predictive, meister-etal-2021-revisiting, wilcox-etal-2023-language, pimentel2023effect}, we operationalize the predictive power as the mean difference in log-likelihood ($\deltaLL$) between the target and the baseline model, \emph{i.e.}, 
\begin{equation}
\small
    \begin{split}
        \deltaLL = \frac{1}{IJ} \left[ \sum_{i=1}^I \sum_{j=1}^J \log \mathcal{M}^t(y_{ij} \mid \mathbf{x}^b_{ij} \oplus\xint_{ij}) \right. \\ - \left. \sum_{i=1}^I \sum_{j=1}^J \log \mathcal{M}^b(y_{ij} \mid \mathbf{x}^b_{ij}) \right], 
    \end{split}
\end{equation}
where $I$ is the number of words and $J$ is the number of subjects.
To avoid overfitting, we perform  10-fold cross validation. A positive $\deltaLL$ indicates a better fit of the target model to the data.

\section{Experiments}
\subsection*{Data}
We employ German reading time data from InDiCo~\citep{haller2023measurement}. 
This corpus contains eye-tracking-while-reading and self-paced-reading data from 61 native German speakers, collected across four experimental sessions, alongside a comprehensive battery of individual psychometric scores in four cognitive domains: cognitive control, working memory, intelligence, and reading fluency.\footnote{For a detailed description of the tests, see Appendix~\ref{sec:appendix:indico}.} For our analyses, we use the standardized scores of 13 psychometric tests. Following previous work \citep[i.a.]{wilcox2023testing}, we employ \textit{first-pass reading time} (FPRT), also referred to as \emph{gaze duration}: the sum of all fixations on a word when fixating it for the first time--as a proxy for processing load. Whereas total fixation duration can incorporate words from the right context due to regressive saccades, FPRT most strongly reflects the initial processing difficulty.\footnote{Contrary to previous work, we do not set the reading time for words that were skipped to zero, but instead exclude them from the analysis. See limitations and caveats in ~\citet{pimentel2023effect} on the influence of skipped words on surprisal estimation.} Given that in our study, we only deploy auto-regressive LMs (cf.\cref{sec:preds}), FPRTs are also more in line with the fact that these models only have access to a word's left context.

\subsection*{Predictors}
\paragraph{Word-level predictors. }\label{sec:preds}
To extract surprisal and contextual entropy estimates, we deploy the German versions of five pretrained transformer-based LMs of different families and sizes, namely GPT-2 base and large~\citep{radford2019language}, Llama 2 7B and 13B~\citep{touvron2023llama}, and Mixtral~\citep{jiang2024mixtral}. 
For details, see Appendix~\ref{sec:appendix:predictor-lms}. Crucially, we only consider auto-regressive LMs, as they most closely align with the incremental nature of human language comprehension~\citep{hale2006uncertainty, rayner2009language}.

Since LMs employ tokenizers which split white-space separated words into sub-word tokens~\citep{sennrich2015neural, song-etal-2021-fast}, word-level surprisal is computed by summing up the surprisal values of the sub-word tokens, which is equivalent to computing the surprisal of the joint distribution of sub-word tokens. Similarly, to obtain the word-level contextual entropy, we use the sum of the sub-word token-level contextual entropy values as proxy for the joint entropy of the sub-word tokens' distributions.\footnote{For details on pooling of surprisal and entropy, see Appendix~\ref{sec:appendix:predictors-pooling}.}

We further include lexical frequency and word length in our analyses since they are known to have an impact on human reading behavior. Lemma frequencies were extracted from dlexDB~\citep{heister2011dlexdb}, based on the reference corpus underlying the Digital Dictionary of the German Language \citep[DWDS;][]{dwds}. \emph{Word length} is defined as the number of characters including punctuation.
Henceforth, we denote the word-level predictors surprisal $s_i$, contextual entropy $h_i$, log-lemma frequency $f_i$, and word length $l_i$ for a word $i$.

\begin{figure*}[htb!]
    \centering
    \includegraphics[width=\linewidth]{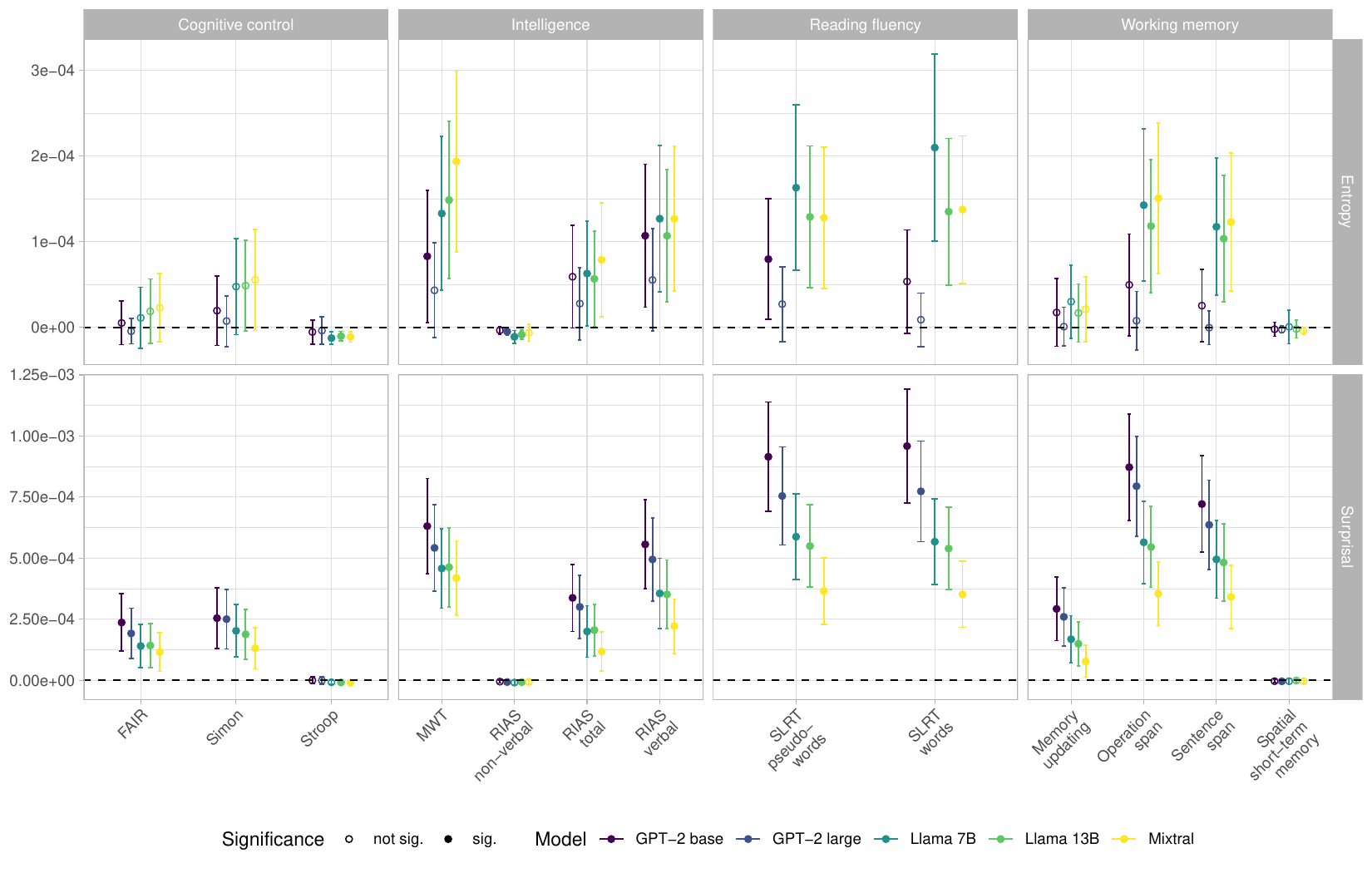}
    \caption{$\deltaLL$ (mean and 95\% CI) for the interactions between psychometric scores and model surprisal or entropy as additional predictors for reading times. Empty dots indicate that the $\deltaLL$ is not significantly different from zero.\looseness=-1} \label{fig:deltall}
\end{figure*}

\paragraph{Psychometric scores. }
The psychometric assessment in InDiCo includes a total of 13 tests targeting different cognitive domains such as verbal and non-verbal working memory, cognitive control and intelligence, as well as reading fluency. A list of tests and their abbreviations can be found in Appendix~\ref{sec:appendix:indico}. For all test scores, higher scores originally indicate higher performance except for the \emph{Stroop Reaction Time Effect} (Stroop) and the \emph{Simon Reaction Time Effect} (Simon). In order to facilitate interpretability, we take the negative values of these scores such that a high value indicates high cognitive control. We standardize all scores in order to facilitate comparisons between tests. We denote the score of a given psychometric test $c$ for subject $j$.

\subsection{Baseline analyses (H\textsubscript{B})}
To corroborate results from previous work, we first assess the predictive power of entropy and surprisal in general, not taking into account individual psychometric scores. 
We define a baseline model $\mathcal{M}^b_0$ with predictors $\mathbf{x}^{b_0}_{i}$ including the word-level predictors {word length} $l_i$, {log-lemma frequency} $f_i$, a global intercept $\beta_0$, and an additional random by-subject intercept $\beta_{0j}$, \emph{i.e.,}
\begin{equation} \label{eq:lmm:b}
\small
\begin{split} 
    \mathcal{M}^b_0: y_{ij} & \sim  \beta_0 + \beta_{0j} + ~ \beta_{1} ~ l_{i} + \beta_{2} ~f_{i},
\end{split}
\end{equation}
where $y_{ij}$ refers to the log-transformed first-pass reading time\footnote{First-pass reading time denotes the sum of all fixation durations on a given word during its first pass (i.e., when reading it for the first time).} of subject $j$ for the $i^{\mathrm{th}}$ word in the stimulus corpus across all texts and following a log-normal distribution.
The target models $\mathcal{M}^{t_{s}}_{0}$ and $\mathcal{M}^{t_{h}}_{0}$ solely include an additional surprisal or entropy term, \ie $s_i$ or $h_i$.

\paragraph{\emph{Results.}}
As depicted in Figure~\ref{fig:baseline}, surprisal and contextual entropy exhibit significant \emph{predictive power} (PP), albeit consistently lower for entropy. For GPT-2 base and large, adding both surprisal and contextual entropy as predictors increases the PP; for the other models, the combined version yields the same PP as using surprisal alone. Across models, GPT-2 base has the highest PP, with PP decreasing as model size increases.


\begin{table*}[htb!]
  \centering
  \small
  \begin{adjustbox}{max width=\textwidth}
  \begin{tabular}{ cllrrrrr }
   \toprule
 & \multirow{2}{*}{{\bf Cognitive domain}} & \multirow{2}{*}{{\bf Test}}&   \multicolumn{5}{c}{Effect size of interaction term} \\
     & & & \multicolumn{1}{c}{GPT-2 \textit{base}} &\multicolumn{1}{c}{GPT-2 \textit{large}} & \multicolumn{1}{c}{Llama-2 7B}  &\multicolumn{1}{c}{Llama-2 13B} &  \multicolumn{1}{c}{Mixtral} \\
     \hline\\[-1.5ex]
\parbox[t]{2mm}{\multirow{13}{*}{\rotatebox[origin=c]{90}{Entropy \phantom{x}}}} & 
\multirow{3}{*}{{ Cognitive control}} &  FAIR & $-0.002 \db{0.001}^{\dag}$ & $-0.001 \db{0.001}^{\dag}$ & $-0.003 \db{0.001}^{\dag}$ & $-0.003 \db{0.001}^{\dag}$ & $-0.003 \db{0.001}^{\dag}$ \\
& & Simon & $0.003 \db{0.001}^{\dag}$ & $0.002 \db{0.001}^{\dag}$ & $0.005 \db{0.001}^{\dag}$ & $0.004 \db{0.001}^{\dag}$ & $0.005 \db{0.001}^{\dag}$ \\
& & Stroop & $-0.001 \db{0.001}^{\dag}$ & $-0.001 \db{0.001}^{\dag}$ & \cellcolor{green!10}$0 \db{0.001}$ & \cellcolor{green!10}$0 \db{0.001}$ & \cellcolor{green!10}$0 \db{0.001}$ \\
& \multirow{4}{*}{{Intelligence}} & MWT & \cellcolor{red!10}$-0.006 \db{0.001}$ & $-0.005 \db{0.001}^{\dag}$ & \cellcolor{red!10}$-0.008 \db{0.001}$ & \cellcolor{red!10}$-0.008 \db{0.001}$ & \cellcolor{red!10}$-0.009 \db{0.001}$ \\
& &RIAS non-verbal & $0 \db{0.001}^{\dag}$ & \cellcolor{green!10}$0 \db{0.001}$ & \cellcolor{green!10}$0 \db{0.001}$ & \cellcolor{green!10}$0 \db{0.001}$ & $-0.001 \db{0.001}^{\dag}$ \\
& & RIAS total & $-0.005 \db{0.001}^{\dag}$ & $-0.004 \db{0.001}^{\dag}$ & $-0.005 \db{0.001}^{\dag}$ & $-0.005 \db{0.001}^{\dag}$ & \cellcolor{red!10}$-0.006 \db{0.001}$ \\
& & RIAS verbal & \cellcolor{red!10}$-0.007 \db{0.001}$ & \cellcolor{red!10}$-0.005 \db{0.001}$ & \cellcolor{red!10}$-0.007 \db{0.001}$ & \cellcolor{red!10}$-0.007 \db{0.001}$ & \cellcolor{red!10}$-0.007 \db{0.001}$ \\
& \multirow{2}{*}{{Reading fluency}} & SLRT pseudo-words & \cellcolor{red!10}$-0.006 \db{0.001}$ & $-0.004 \db{0.001}^{\dag}$ & \cellcolor{red!10}$-0.008 \db{0.001}$ & \cellcolor{red!10}$-0.007 \db{0.001}$ & \cellcolor{red!10}$-0.007 \db{0.001}$ \\
& & SLRT words & $-0.005 \db{0.001}^{\dag}$ & $-0.003 \db{0.001}^{\dag}$ & \cellcolor{red!10}$-0.009 \db{0.001}$ & \cellcolor{red!10}$-0.007 \db{0.001}$ & \cellcolor{red!10}$-0.007 \db{0.001}$ \\
 & \multirow{4}{*}{{Working memory}} & Memory updating & $-0.003 \db{0.001}^{\dag}$ & $-0.002 \db{0.001}^{\dag}$ & $-0.004 \db{0.001}^{\dag}$ & $-0.003 \db{0.001}^{\dag}$ & $-0.003 \db{0.001}^{\dag}$ \\
& & Operation span & $-0.005 \db{0.001}^{\dag}$ & $-0.003 \db{0.001}^{\dag}$ & \cellcolor{red!10}$-0.008 \db{0.001}$ & \cellcolor{red!10}$-0.007 \db{0.001}$ & \cellcolor{red!10}$-0.008 \db{0.001}$ \\
& & Sentence span & $-0.003 \db{0.001}^{\dag}$ & $-0.002 \db{0.001}^{\dag}$ & \cellcolor{red!10}$-0.007 \db{0.001}$ & \cellcolor{red!10}$-0.006 \db{0.001}$ & \cellcolor{red!10}$-0.007 \db{0.001}$ \\
& & Spatial short-term memory & $-0.001 \db{0.001}^{\dag}$ & $0 \db{0.001}^{\dag}$ & $0.002 \db{0.001}^{\dag}$ & $0.001 \db{0.001}^{\dag}$ & $0 \db{0.001}^{\dag}$ \\
    \midrule
\parbox[t]{2mm}{\multirow{13}{*}{\rotatebox[origin=c]{90}{Surprisal \phantom{x}}}} &    
\multirow{3}{*}{{Cognitive control}} & FAIR & \cellcolor{red!10}$-0.01 \db{0.001}$ & \cellcolor{red!10}$-0.009 \db{0.001}$ & \cellcolor{red!10}$-0.008 \db{0.001}$ & \cellcolor{red!10}$-0.008 \db{0.001}$ & \cellcolor{red!10}$-0.007 \db{0.001}$ \\
& & Simon & \cellcolor{green!10}$0.01 \db{0.001}$ & \cellcolor{green!10}$0.01 \db{0.001}$ & \cellcolor{green!10}$0.009 \db{0.001}$ & \cellcolor{green!10}$0.008 \db{0.001}$ & \cellcolor{green!10}$0.007 \db{0.001}$ \\
& & Stroop & $-0.001 \db{0.001}^{\dag}$ & $-0.001 \db{0.001}^{\dag}$ & $-0.001 \db{0.001}^{\dag}$ & \cellcolor{green!10}$0 \db{0.001}$ & \cellcolor{green!10}$0 \db{0.001}$ \\
& \multirow{4}{*}{{Intelligence}} & MWT & \cellcolor{red!10}$-0.016 \db{0.001}$ & \cellcolor{red!10}$-0.015 \db{0.001}$ & \cellcolor{red!10}$-0.014 \db{0.001}$ & \cellcolor{red!10}$-0.014 \db{0.001}$ & \cellcolor{red!10}$-0.014 \db{0.001}$ \\
& & RIAS non-verbal & $0 \db{0.001}^{\dag}$ & \cellcolor{green!10}$0 \db{0.001}$ & $0.001 \db{0.001}^{\dag}$ & \cellcolor{green!10}$0 \db{0.001}$ & $0.001 \db{0.001}^{\dag}$ \\
& & RIAS total & \cellcolor{red!10}$-0.011 \db{0.001}$ & \cellcolor{red!10}$-0.011 \db{0.001}$ & \cellcolor{red!10}$-0.009 \db{0.001}$ & \cellcolor{red!10}$-0.009 \db{0.001}$ & \cellcolor{red!10}$-0.007 \db{0.001}$ \\
& & RIAS verbal & \cellcolor{red!10}$-0.015 \db{0.001}$ & \cellcolor{red!10}$-0.014 \db{0.001}$ & \cellcolor{red!10}$-0.012 \db{0.001}$ & \cellcolor{red!10}$-0.012 \db{0.001}$ & \cellcolor{red!10}$-0.01 \db{0.001}$ \\
& \multirow{2}{*}{{ Reading fluency}} & SLRT pseudo-words & \cellcolor{red!10}$-0.018 \db{0.001}$ & \cellcolor{red!10}$-0.017 \db{0.001}$ & \cellcolor{red!10}$-0.015 \db{0.001}$ & \cellcolor{red!10}$-0.014 \db{0.001}$ & \cellcolor{red!10}$-0.012 \db{0.001}$ \\
& & SLRT words & \cellcolor{red!10}$-0.019 \db{0.001}$ & \cellcolor{red!10}$-0.017 \db{0.001}$ & \cellcolor{red!10}$-0.015 \db{0.001}$ & \cellcolor{red!10}$-0.014 \db{0.001}$ & \cellcolor{red!10}$-0.012 \db{0.001}$ \\
&  \multirow{4}{*}{{Working memory}} & Memory updating & \cellcolor{red!10}$-0.011 \db{0.001}$ & \cellcolor{red!10}$-0.01 \db{0.001}$ & \cellcolor{red!10}$-0.008 \db{0.001}$ & \cellcolor{red!10}$-0.008 \db{0.001}$ & \cellcolor{red!10}$-0.006 \db{0.001}$ \\
& & Operation span & \cellcolor{red!10}$-0.018 \db{0.001}$ & \cellcolor{red!10}$-0.018 \db{0.001}$ & \cellcolor{red!10}$-0.015 \db{0.001}$ & \cellcolor{red!10}$-0.015 \db{0.001}$ & \cellcolor{red!10}$-0.012 \db{0.001}$ \\
& & Sentence span & \cellcolor{red!10}$-0.016 \db{0.001}$ & \cellcolor{red!10}$-0.015 \db{0.001}$ & \cellcolor{red!10}$-0.014 \db{0.001}$ & \cellcolor{red!10}$-0.013 \db{0.001}$ & \cellcolor{red!10}$-0.012 \db{0.001}$ \\
& & Spatial short-term memory & $0 \db{0.001}^{\dag}$ & \cellcolor{green!10}$0 \db{0.001}$ & $0.001 \db{0.001}^{\dag}$ & $0.001 \db{0.001}^{\dag}$ & $0.001 \db{0.001}^{\dag}$ \\
\bottomrule
  \end{tabular} 
  \end{adjustbox}
  \caption{Effect sizes of interaction terms $\pm$ standard error between \textbf{entropy} (top) / \textbf{surprisal} (bottom) and psychometric test scores.$^{\dag}$ indicates that the inclusion of the interaction term did not lead to a significant increase or decrease in $\deltaLL$ (see Figure~\ref{fig:deltall}). \looseness=-1}
  \label{tab:res-effects}
\end{table*}

\subsection{Assessing the predictive power of interactions between surprisal/entropy and psychometric scores (H\textsubscript{1})} \label{sec:methods:deltall}
To examine whether an interaction between cognitive scores and surprisal or entropy leads to an increase in predictive power on reading times, we define a baseline model $\mathcal{M}^b_1$ with predictors $\mathbf{x}^{b_1}_{ij}$ including the word-level predictors $l_i$, $f_i$, $s_i$, $h_i$, and the subject-level predictor $c_j$ denoting the test score of a specific psychometric test (e.g., \emph{word-reading fluency}) obtained for subject $j$, and again a by-subject intercept $\beta_{0j}$, \emph{i.e.,}
\begin{equation} \label{eq:lmm:h1b}
\small
\begin{split} 
    \mathcal{M}^b_1: y_{ij} & \sim  \beta_0 + \beta_{0j} + ~ \beta_{1} ~ l_{i} + \beta_{2} ~f_{i} + \\ &\quad \beta_{3} ~s_{i} +  \beta_4~h_i + \beta_{5}~c_j
\end{split}
\end{equation}

To assess whether allowing surprisal or entropy to be modulated by specific cognitive profiles---operationalized in terms of the individual psychometric measures---improves the prediction of reading time, we define target models $\mathcal{M}^{t_{s}}_{1}$ and $\mathcal{M}^{t_{h}}_{1}$ that include an additional interaction term between either surprisal or entropy and a given psychometric score $c_j$ (e.g., \emph{word-reading fluency score}) obtained for subject $j$, $x^{q_1}_{ij}\in \{s_i\cdot c_j, h_i\cdot c_j\}$:
\begin{equation}
\small
\begin{split} \label{eq:lmm:t}
    \mathcal{M}^t_1: y_{ij} & \sim \beta_0 + \beta_{0j} + ~ \beta_{1} ~ l_{i} + \beta_{2} ~f_{i} + \\ &\quad \beta_3~s_i +  \beta_{4} ~h_{i} +\beta_{5} ~ c_{j} +\beta_{6} ~ x^{q_1}_{ij}
\end{split}
\end{equation}
A positive $\deltaLL$ between the target and the baseline model indicates that including the participant's score of a given psychometric test improves the prediction on the held-out test data. We run paired permutation tests using the R library \texttt{broman} to establish whether a given  $\deltaLL$ is significantly different from $0$ at $\alpha=.05$.



\paragraph{\emph{Results.}}
Figure \ref{fig:deltall} shows the $\deltaLL$ across all psychometric tests and models. 
Overall, we see that the interaction terms between surprisal/entropy and most psychometric scores lead to significant increases in PP, except for Stroop, non-verbal RIAS and spatial short-term memory. Notably, PP is not significant (or extremely small) for these three scores across all models. Additionally, there are notable differences among different cognitive domains with respect to predictive power: modulating surprisal with scores targeting reading fluency or working-memory span yields the highest predictive power, followed by verbal intelligence scores. Scores targeting cognitive control show the lowest PP. 
The pattern observed in Figure~\ref{fig:baseline} showing that increasing model size is associated with decreasing predictive power is visible here as well, but only for surprisal, not for entropy. Overall, interactions with surprisal extracted from the GPT-2 family have the highest PP. Conversely, interactions with GPT-2 based entropy have the lowest PP.


\begin{figure*}[htb!]
    \centering
    \includegraphics[width=\linewidth]{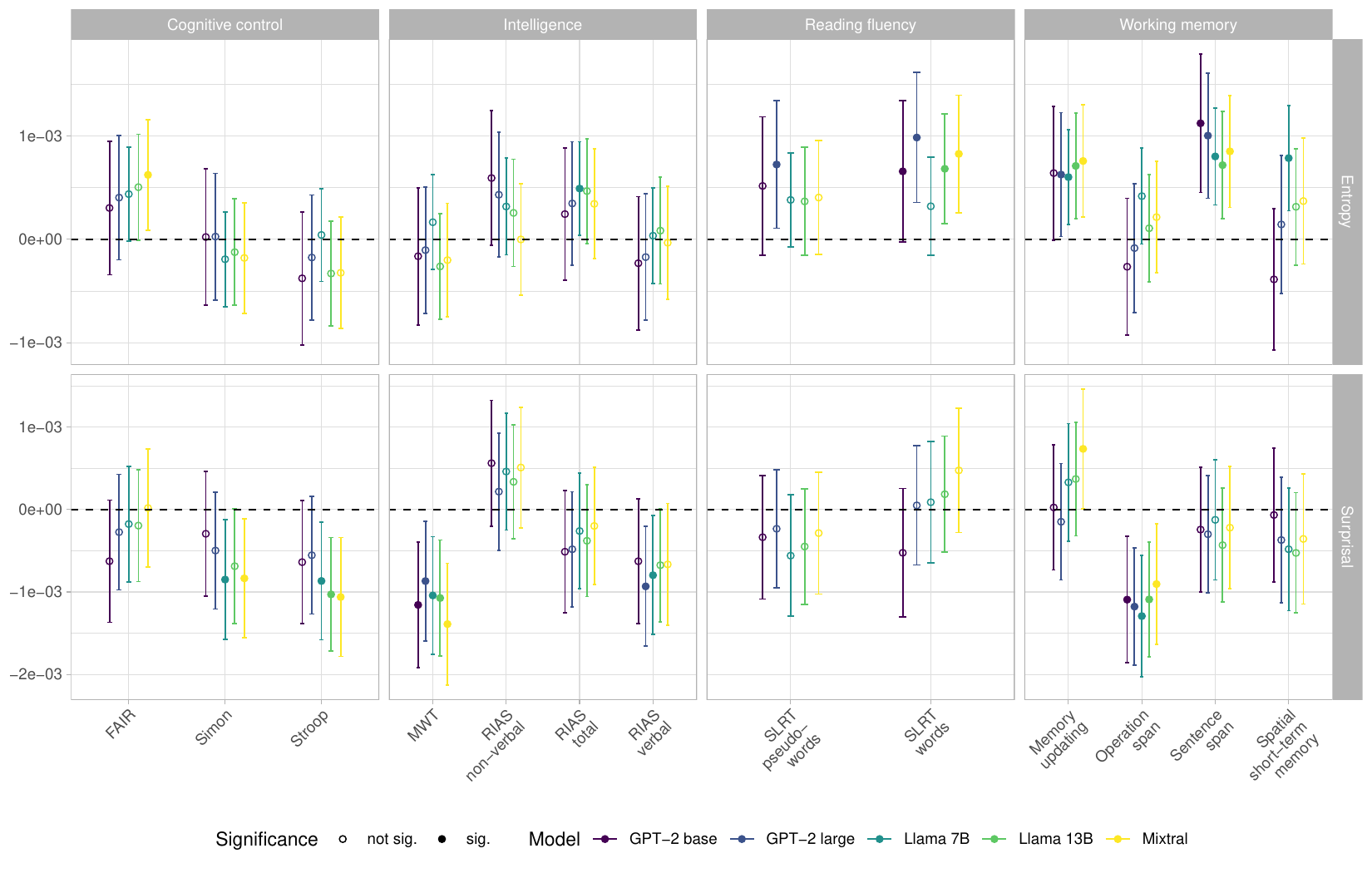}
    \caption{Difference in PP ($\pp$) (mean and 95\% CI) of surprisal and contextual entropy for reading times. Positive $\pp$ indicate higher PP for high-performing individuals; negative $\pp$ higher PP for low-performing individuals. Empty dots indicate that the $\pp$ is not significantly different from zero. \looseness=-1} \label{fig:pg}
\end{figure*}

\subsection{Assessing the magnitude of the interaction term coefficients (H\textsubscript{2})}
\label{sec:experiments:h2}
To determine how specifically the surprisal and entropy effects are modulated by the psychometric
scores, we run the target models $\mathcal{M}^{t_{s}}_{1}$ and $\mathcal{M}^{t_{h}}_{1}$ on the entire dataset and examine the effect sizes (coefficients) of the interaction term between the scores and the surprisal and entropy estimates, $\beta_6$. 
The coefficient of the interaction term indicates to what degree the fixed effect surprisal or entropy term is adjusted, relative to a subject's individual psychometric score, or, in other words, if individuals with a given cognitive profile are more sensitive to predictability effects. For instance, a positive coefficient for predictor $c_j\cdot s_i$ indicates that subjects with a higher score exhibit a stronger effect of surprisal, i.e.~are more sensitive to a word's predictability, while a negative coefficient implies that subjects with a lower score exhibit a higher surprisal effect.

\paragraph{\emph{Results.}}
We present the effect sizes of the interaction between scores and predictability measures in Table~\ref{tab:res-effects}. First, we notice that for a given psychometric test, all models consistently modulate surprisal and entropy effects in the same direction. For most psychometric tests, higher scores result in a reduction of surprisal and entropy effects, indicated by the negative interaction term coefficients. In these cases, individuals with higher scores show less sensitivity to a word's predictability (measured in terms of surprisal or entropy). This holds true across all tests, the only exception being the Simon test, providing a measure of non-verbal inhibitory cognitive control. Here, high-performing individuals exhibit larger surprisal effects. Positive coefficients are also found for the Stroop task and the non-verbal part of the RIAS (intelligence), although they are extremely small. 
 
\subsection{Assessing the difference in predictive power between cognitive profiles (H\textsubscript{3})}
\label{sec:experiments:h3}
Finally, we investigate whether there are differences in the predictive power of LM surprisal and entropy for reading times obtained from individuals with different cognitive profiles. In other words, we ask the question what type of psycholinguistic subject a given language model emulates. To do so, we split the reading time data into subsets of high-performing ($\uparrow$) and low-performing individuals ($\downarrow$) at the median of each score. Then, for each group, we compute the $\deltaLL$ between the baseline model $\mathcal{M}^b_3$ only including word length and lexical frequency as predictors and the target model $\mathcal{M}^t_3$ with an additional predictor of interest $x^{q_3}_{i}\in \{s_i, h_i\}$, i.e.,~either surprisal or entropy. The individual ${\deltaLL}_{\downarrow}$ and $\deltaLL_{\uparrow}$ indicate the predictive power of surprisal and entropy for each group separately. In order to answer which group exhibited a higher relative gain in PP, we assess the difference in predictive power $\pp\defeq\deltaLL_{\uparrow}-\deltaLL_{\downarrow}$. If a given LM is calibrated towards the cognitive profile of the high-performing individuals, we expect a positive $\pp$; and the $\pp$ is negative if the LM is calibrated towards the low-performing cognitive profile. 

\paragraph{\emph{Results.}}
Figure~\ref{fig:pg} presents the differences (mean and 95\% CI) in predictive power ($\pp$) of surprisal or entropy between two groups that performed above or below the median, respectively, in a given psychometric test. $\pp > 0$ indicates higher PP for the high-performing group, $\pp < 0$ indicates higher PP for the low-performing group. 

First, looking at the results for entropy, we note that across all models, entropy predicts the RTs of individuals among the high-performing groups in the memory-updating and operation-span tests significantly better. Regarding reading fluency, readers with high word-reading scores were significantly better predicted from entropy estimated via GPT-2 large, Llama 2 13B and Mixtral. 

For surprisal, we find that across all models, RT predictions are significantly better for the the low-performing group in the operation span test as well as the vocabulary size test MWT. Moreover, surprisal extracted from GPT-2 large and Llama 2 7B leads to significant gains in PP for the low-performing group in the RIAS test, which like MWT assesses verbal intelligence. Finally, surprisal estimated from Llama 2 7B and Mixtral showed significantly higher PP for the group of individuals with lower scores in the verbal and non-verbal cognitive control tests (Stroop, Simon).

\section{Discussion}

Most studies on the predictive power of surprisal and entropy on reading times have been conducted on the group-level. Although individual-level effects may have been taken into account in terms of random slopes, these effects have not been associated with different cognitive profiles. Our findings suggest that (1) incorporating information on individuals' cognitive capacities and allowing them to modulate the magnitude of surprisal and entropy effects can increase the predictive power of these predictability measures, (2) individuals exhibit surprisal and entropy effects relative to certain cognitive capacities, and (3) some language models exhibit higher predictive power of reading times for groups of individuals associated with a certain cognitive profile.

\subsection{Implications for the cognitive mechanisms of language processing}
\paragraph{Fluent readers exhibit lower surprisal effects.\footnote{As many results for entropy are not significant or less clear, the focus of the discussion will lie on surprisal.}}
Our results show that the predictive power of the interaction terms between surprisal and reading fluency and to some extent entropy and reading fluency are particularly high compared to the interaction terms including other psychometric tests, as depicted in Figure~\ref{fig:deltall}. Including the interaction term between reading fluency scores and surprisal improves the predictions on reading times for all language models. The negative coefficient (Table~\ref{tab:res-effects}) can be interpreted from two perspectives. From the participants' perspective, it underlines that individuals with high reading fluency exhibit lower surprisal effects. These results might indicate that less fluent readers rely more on predictive processing, hence their reading is easily interrupted by less predictable continuations, leading to longer reading times. Experienced readers, on the other hand, might be more trained to integrate unexpected material effortlessly. From the models' perspective, on the other hand, it means that LMs overestimate the surprisal effect exhibited by highly fluent readers. Similar arguments can be made for the verbal intelligence test (RIAS-verbal), which is correlated with reading fluency (cf.~Figure~\ref{fig:corrmat}).

\paragraph{More accurate predictive processing for individuals with high working memory span. }
The results regarding the interaction terms between surprisal and working memory test scores are more difficult to contextualize. The span tests in particular (operation span, sentence span) lead to substantial increases in PP. Moreover, when assessing the magnitude of their interaction terms, we find that individuals with higher scores in both tests show lower surprisal effects as shown in Table~\ref{tab:res-effects}. At first glance, this intuitively makes sense, as high working memory can be associated with the capability to hold competing continuations in memory, including less likely ones that, in the high-surprisal situation, turn out to be the actual continuation. However, conversely, \citet{o2013interaction} found in an ERP study that individuals with high operation span show stronger P600 effects, an event-related potential that is typically associated with syntactic repair or reanalysis. This would suggest that individuals with higher operation span exhibit a stronger garden path effect. However, garden paths are an extreme case of very high surprisal where different processing mechanisms might be at work such as re-analysis processes. Future experiments in minimal-pair settings will be needed to examine the connection between working memory capacity and surprisal effects more closely.

\paragraph{Attentiveness and inhibitory cognitive control may impact predictive processing differently. }
Next, we discuss the interaction term between surprisal and measures from tests targeting cognitive control. The directions of the interaction coefficients indicate that individuals with higher FAIR-scores (attention and concentration) exhibit weaker surprisal effects. Although this finding would be in line with the fact that general-purpose cognitive control mechanisms are required in the revision after linguistic misanalyses~\citep{fedorenko2014role}, \citet{Vuong2014} has shown that the time taken to revise a garden path is correlated only with verbal Stroop reaction time effects, but not with reaction time effects from its non-verbal counterpart (Simon). Since the FAIR scores showed fairly strong correlations with working-memory scores but not the other cognitive control tests (cf. Fig.~\ref{fig:corrmat}), it is possible that the results obtained for FAIR might be more related to working-memory principles.
Secondly, the results also showed that individuals performing well in the Simon test (inhibitory non-verbal cognitive control) exhibit stronger surprisal effects. As mentioned before, \citet{Vuong2014} showed that the Simon task is likely not associated with mechanisms related to linguistic repair. The weaker surprisal effects for low-performing individuals in the Simon task is more likely associated with the tendency of participants with lower control to skip revising misinterpretations entirely, i.e., to rely on good-enough processing~\citep{ferreira2002good}. 
\subsection{Cognitive profiles of language models}
Finally, regarding the group analyses, the results presented in Figure~\ref{fig:pg} revealed that surprisal estimates across all tested models predicted RTs better for the group of individuals with low verbal intelligence scores, measured with two largely complementary tests: one that assesses word knowledge (MWT-B), and one that assesses verbal logical thinking via question answering and sentence completion (RIAS-verbal). At first glance, this result is surprising since a language model has been exposed to billions of tokens, and therefore, one might expect that it emulates a psycholinguistic subject with high verbal intelligence. However, a language model’s predictions are always relative, i.e., even if it has seen infrequent words, it will still have a preference in terms of likelihood for the more regular, frequent continuation. Individuals with high verbal intelligence do not struggle with such contexts since they are very familiar even with uncommon terminology.

Additionally, we found that the PP of entropy is significantly higher for individuals with high working memory capacities, measured via memory updating and sentence span. This result suggests that uncertainty measures about upcoming material exhibited by LMs are more in line with the way high-working memory individuals process language, potentially driven by taking into account longer contexts, or keeping track of relevant long dependencies.

Even though most results from all three experiments are consistent within and across different LM families, there are exceptions. For instance, entropy estimated from GPT-2 large showed the strongest increase in PP for the high reading-fluency \emph{word reading} group (Figure \ref{fig:pg}). For the high reading-fluency \emph{pseudo-word reading} group, it even represents the only measure with a significant increase in PP. Taking into account that GPT-2 base and large showed a similar baseline PP (Figure~\ref{fig:baseline}), this suggests that entropy extracted from GPT-2 large is a better proxy of processing effort for readers with lower verbal intelligence than entropy estimated with GPT-2 base. This illustrates that the choice of LM to estimate predictability measures is crucial for downstream analyses in psycholinguistic studies or NLP applications, especially when working with specific target groups. In such settings, it might be worthwhile considering a model that is less biased, or, in other words, whose predictability measures are well-aligned with the target group at hand as it will most likely lead to more accurate results.

While this study aimed at uncovering model-internal biases, it might be worthwhile to, in turn, extend the investigation on whether text \emph{produced} by a given LM is biased towards being processed more easily by individuals with specific cognitive characteristics. This is particularly important for tasks such as text summarization or simplification that might need to be tailored to specific groups.

\section{Conclusion}
To date, most investigations on predictability effects have been conducted on the group-level, disregarding individual differences, assuming that the predictive power of next-word predictability metrics such as surprisal or entropy on human reading times is uniform across cognitive profiles. In this work, we have shown that indeed, LMs do exhibit systematic biases towards readers of certain cognitive profiles. This illustrates the usefulness of incorporating individual-level information within the study of LM interpretability and language modelling in general.
\clearpage

\section*{Limitations}
\label{sec:limitations}

Splitting the subjects at the median of their scores obtained in the respective cognitive tests is a straightforward way to split them into high- and low-performing groups and avoids the problem of class-imbalance. However, this kind of split might not group participants whose scores are distributed more narrowly. For future work, it might be sensible to utilize more sophisticated clustering approaches to obtain more cognitively homogeneous groups.
    
Moreover, recent work has shown that test-retest reliability of individual surprisal effects are low, \ie a surprisal effect for the same individual might vary on different days~\citep{haller2023measurement}, depending on numerous factors such as wakefulness, motivation, but also random fluctuations. If this is true, we have to assume the predictive power with one and the same LM for a given subject, representing a cognitive profile, might be different depending on the subject's condition on that particular day. However, using the Indico data, this factor is controlled to some degree since it combines data from temporally separate sessions. That way, even if the surprisal effect does depend on external factors, merging the data from several sessions ensures more robust estimates of each subject’s true surprisal effect. While it might still be a limitation, it is less so than for other conventional datasets. Finally, although InDiCo represents a fairly diverse sample in terms of age and gender, many participants have an academic background (see also \citet{reich-etal-2024-reading-equal} on the necessity of including diverse populations in analyses of reading data).


\section*{Ethics Statement}
\label{sec:ethics-statement}

Working with human data requires careful ethical considerations. The \emph{Individual Differences Corpus}~\citep[InDiCo;][]{haller2023measurement} utilized for this study follows the Helsinki Declaration~\citep{world2013world}.

\section*{Acknowledgements}
This work was partially funded by the Swiss National Science Foundation  under grant 100015L\_212276/1 (MeRID). We thank David Reich, Nora Hollenstein and Omer Shubi for valuable discussions regarding this work.

\bibliography{anthology,custom}
\bibliographystyle{acl_natbib}

\clearpage

\onecolumn

\appendix

\section{Details on predictors} \label{sec:appendix:predictor-details}

\subsection{Language Models}
\label{sec:appendix:predictor-lms}

We deployed the following German LMs from the Huggingface library~\citep{wolf2019huggingface}:
\begin{itemize}[itemsep=0pt, parsep=0pt]
    \item GPT-2 \textit{base}: \url{https://huggingface.co/benjamin/gerpt2}
    \item GPT-2 \textit{large}: \url{https://huggingface.co/benjamin/gerpt2-large}
    \item Llama 2 7B: \url{https://huggingface.co/LeoLM/leo-hessianai-7b}
    \item Llama 2 13B: \url{https://huggingface.co/LeoLM/leo-hessianai-13b}
    \item Mixtral: \url{https://huggingface.co/mistralai/Mixtral-8x7B-v0.1}
\end{itemize}

\subsection{Pooling of surprisal and contextual entropy to word level}
\label{sec:appendix:predictors-pooling}
We compute word-level surprisal by summing up the surprisal values of the individual sub-word tokens. Given $k$ sub-word tokens $u_n, u_{n+1}, \dots, u_{n+k}$ belonging to the same word token, the word token's surprisal is computed as 
\begin{align*}
    s(u_n, u_{n+1}, \dots, u_{n+k})
    & = - \log p(u_n, u_{n+1}, \dots, u_{n+k} \mid \mathbf{u}_{<n}) \\
    & = \begin{aligned}
        - \log \left[ p(u_n \mid \mathbf{u}_{<n}) p(u_{n+1} \mid \mathbf{u}_{< n+1}) \dots p(u_{n+k} \mid \mathbf{u}_{< n+k}) \right] 
    \end{aligned} \\
    & = \begin{aligned}
        - \log p(u_n \mid \mathbf{u}_n) + - \log p(u_{n+1} \mid \mathbf{u}_{< n+1}) + \dots + - \log p(u_{n+k} \mid \mathbf{u}_{< n+k}),
    \end{aligned}
\end{align*}

which shows that summing up sub-word token surprisal values is equivalent to computing the surprisal of the joint distribution of the sub-word tokens.

As regards entropy, we use the sum of the sub-word token-level contextual entropies as proxy for the joint entropy of the sub-word tokens' distribution. Given $k$ $\Bar{\Sigma}$-valued random variables $U_n, U_{n+1}, \dots, U_{n+k}$ belonging to the same word token, their joint entropy is defined as:
\begin{equation*}
\small
\begin{split}
    \mathrm{H}(U_n, U_{n+1}, \dots, U_{n+k}) \defeq -\sum_{u_n \in \Bar{\Sigma}} \sum_{u_{n+1} \in \Bar{\Sigma}} \dots \sum_{u_{n+k} \in \Bar{\Sigma}} P(u_n, u_{n+1}, \dots, u_{n+1}) \log_2 \left[ P(u_n, u_{n+1}, \dots, u_{n+1}) \right].
\end{split}
\end{equation*}
However, depending on the tokenizer, the cardinality of $\Bar{\Sigma}$ could be over 50,000, which makes the computation of the joint entropy computationally unfeasible. Instead, we use the sum of the individual entropies as proxy. This is only a proxy, since 
\begin{equation*}
    \mathrm{H}(U_n, U_{n+1}, \dots, U_{n+k}) \leq \mathrm{H}(U_n) + \mathrm{H}(U_{n+1}) + \dots + \mathrm{H}(U_{n+k}).
\end{equation*}
This inequality is an equality iff $U_n, U_{n+1}, \dots, U_{n+k}$ are statistically independent. Since this is not the case here, the sum of the sub-word token-level entropies is used as an upper bound.


\section{Individual Differences Corpus (InDiCo)} \label{sec:appendix:indico}
\begin{figure}
    \centering
    \includegraphics[width=.9\linewidth]{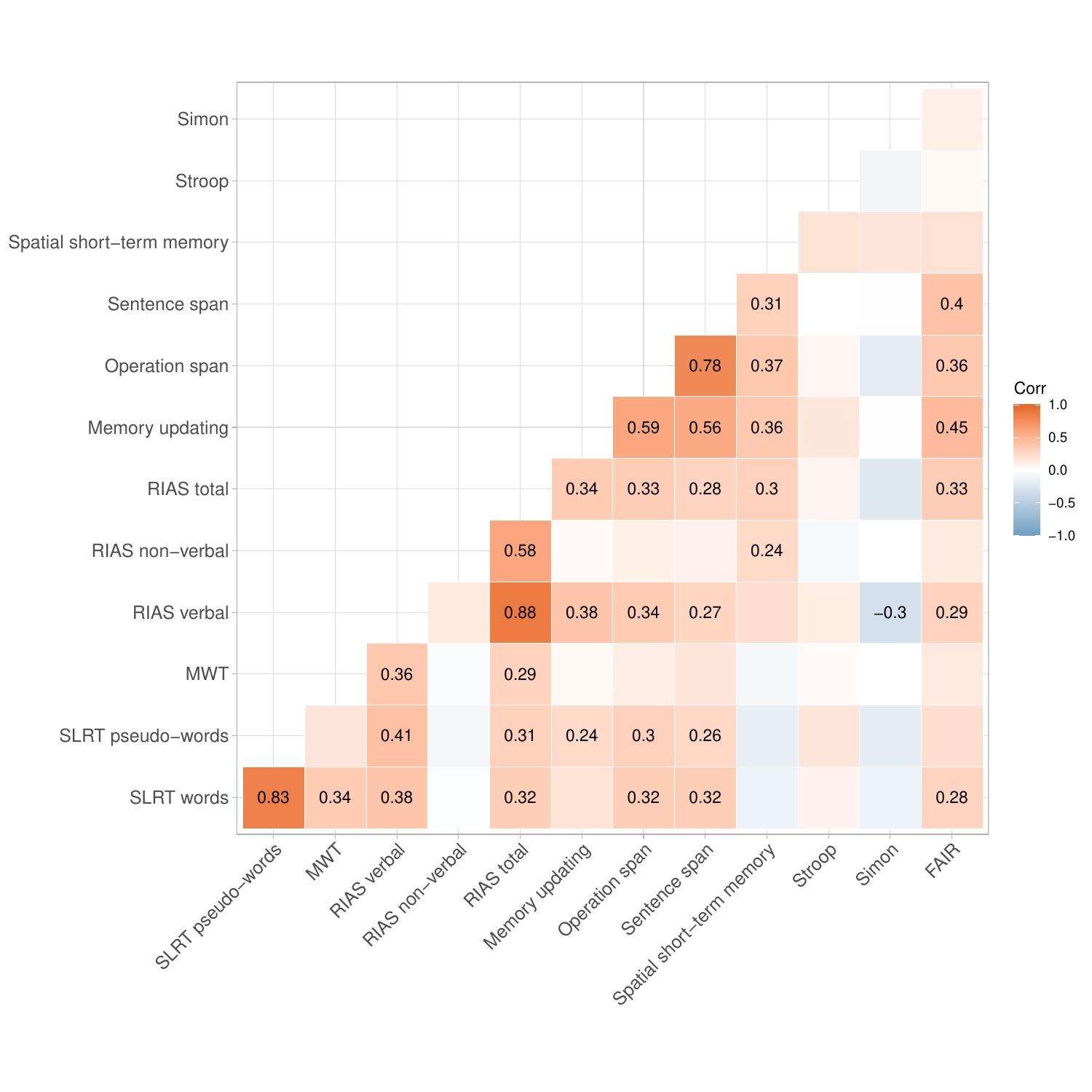}
    \caption{Correlations between scores of all psychometric tests. Red cells indicate positive correlation coefficients, blue cells negative correlation coefficients. Significant coefficients are displayed, blank cells indicate that the correlation was not significant with $\alpha=.05$.}
    \label{fig:corrmat}
\end{figure}

We provide abbreviations and a brief summary of all psychometric tests in Table \ref{tab:tests}. More details can be found in \citet{haller2023measurement}. A correlation matrix between all tests can be found in Figure~\ref{fig:corrmat}. We can see strong correlations between many tests, in particular for the ones of the same psychological construct.

\begin{sidewaystable}
\renewcommand{\arraystretch}{1}
\small
\begin{tabular}{>{\raggedright}p{.02\textwidth} 
                >{\raggedright\arraybackslash}p{.14\textwidth}>{\raggedright}p{.1\textwidth}
                >{\raggedright\arraybackslash}p{.6\textwidth}}
\toprule
& \textbf{Test / Measure} & \textbf{Construct}            & \textbf{Description}   \\ 
\midrule
\parbox[t]{2mm}{\multirow{3}{*}{\rotatebox[origin=c]{90}{Cognitive control \phantom{x}}}}               & Stroop: reaction time effect           & Verbal inhibitory cognitive control     & Participants had to react (choose between congruent and incongruent) for color words whose font color either matched the content (congruent) or not (incongruent). Reaction time and accuracy were measured. \\
& Simon: reaction time effect           & Non-verbal inhibitory cognitive control & Non-verbal equivalent to the Stroop Task where participants had to react (choose between congruent and incongruent) to arrows pointing to the right or left, shown either on the left or right side of the screen.                                        \\
& FAIR: K score (total score)               & Non-verbal cognitive control/attention  & Participants had to find and mark target symbols (e.g., dice with 2 eyes among many other dice) on a page within a time limit. Measures of attentional performance, attention quality, and attention continuity were derived.                          \\
\cmidrule{2-4}
\parbox[c]{2mm}{\multirow{4}{*}{\rotatebox[origin=c]{90}{Working memory\phantom{xxxxx}}}} & Sentence span                           & Verbal working memory capacity          & Participants had to judge the meaningfulness of sentences and remember letters presented after each sentence for later recall. In the end, they had to repeat all the letters.                                                                             \\
& Operation span               & Non-verbal working memory capacity      & Participants were presented with consonants sequentially. After each consonant, they had to perform mathematical operations before the next consonant appeared. In the end, they had to repeat all consonants.                                             \\
& Memory updating               & Non-verbal working memory capacity      & Participants had to remember an initial set of digits, each presented in a separate frame on the screen, and then update these digits in parallel through arithmetic operations.                                                                           \\
& Spatial short-term memory       & Non-Verbal Working Memory Capacity      & Participants had to memorize the spatial locations of dots in a grid during a learning phase, and then locate them on an empty grid. \\

\cmidrule{2-4}
\parbox[t]{2mm}{\multirow{4}{*}{\rotatebox[origin=c]{90}{Intelligence\phantom{xxxx}}}} & MWT: Percentile rank                                       & Verbal intelligence/ word knowledge     & Participants were presented lists of words, and for each list, they had to decide which of the presented words were real words.                                                                                                                            \\
& RIAS: verbal percentile rank             & Verbal Intelligence                     & This test assessed verbal reasoning, and verbal logical thinking via question-answering and sentence completion.                                                                                                                                           \\
& RIAS: non-verbal percentile rank            & Non-Verbal Intelligence                 & This test assessed non-verbal reasoning and problem-solving tests where participants were presented with sets of images and they had to decide which image was not part of the set. In the other test, they had to identify missing elements in pictures.  \\
& RIAS: total percentile rank            & Intelligence                 & Total intelligence score based on verbal and non-verbal part. \\

\cmidrule{2-4}
\parbox[t]{2mm}{\multirow{2}{*}{\rotatebox[origin=c]{90}{Reading\phantom{x}}}} &
 SLRT: Word reading percentile rank & Reading fluency    &                      Participants read out loud as many words within one minute as possible. \\
 & SLRT: Pseudoword reading percentile rank & Reading fluency    &  Participants read out as many pseudo-words within one minute as possible. \\
\bottomrule
\end{tabular}
  \caption{\small{Psychometric tests conducted with all participants.}}
  \label{tab:tests}
\end{sidewaystable}

\end{document}